\pdfoutput=1

\documentclass[11pt]{article}

\usepackage{acl}

\usepackage{times}
\usepackage{latexsym}

\usepackage[T1]{fontenc}

\usepackage[utf8]{inputenc}

\usepackage{microtype}

\usepackage{inconsolata}

\usepackage{subcaption}
\usepackage{graphicx}
\usepackage{amsfonts} 
\usepackage{amsmath} 
\usepackage{bm}
\usepackage{float}
\usepackage{xcolor}

\DeclareMathOperator*{\diag}{diag}
\DeclareMathOperator*{\LN}{LN}

\title{GADePo: Graph-Assisted Declarative Pooling Transformers for Document-Level Relation Extraction}

\author{
Andrei C. Coman \\
Idiap Research Institute, EPFL \\
\href{mailto:andrei.coman@idiap.ch}{andrei.coman@idiap.ch} \\
\And
Christos Theodoropoulos \\
KU Leuven \\
\href{mailto:christos.theodoropoulos@kuleuven.be}{christos.theodoropoulos@kuleuven.be}
\AND
Marie-Francine Moens \\
KU Leuven \\
\href{mailto:sien.moens@kuleuven.be}{sien.moens@kuleuven.be}
\And
James Henderson \\
Idiap Research Institute \\
\href{mailto:james.henderson@idiap.ch}{james.henderson@idiap.ch}
}

\begin{document}
\maketitle
\begin{abstract}
Document-level relation extraction typically relies on text-based encoders and hand-coded pooling heuristics to aggregate information learned by the encoder. In this paper, we leverage the intrinsic graph processing capabilities of the Transformer model and propose replacing hand-coded pooling methods with new tokens in the input, which are designed to aggregate information via explicit graph relations in the computation of attention weights.
We introduce a joint text-graph Transformer model and a \underline{g}raph-\underline{a}ssisted \underline{de}clarative \underline{po}oling (GADePo) specification of the input, which provides explicit and high-level instructions for information aggregation. GADePo allows the pooling process to be guided by domain-specific knowledge or desired outcomes but still learned by the Transformer, leading to more flexible and customisable pooling strategies.
We evaluate our method across diverse datasets and models and show that our approach yields promising results that are consistently better than those achieved by the hand-coded pooling functions.
\end{abstract}

\section{Introduction}
\label{sec:introduction}

\begin{figure}
    \centering
    \includegraphics[width=\linewidth]{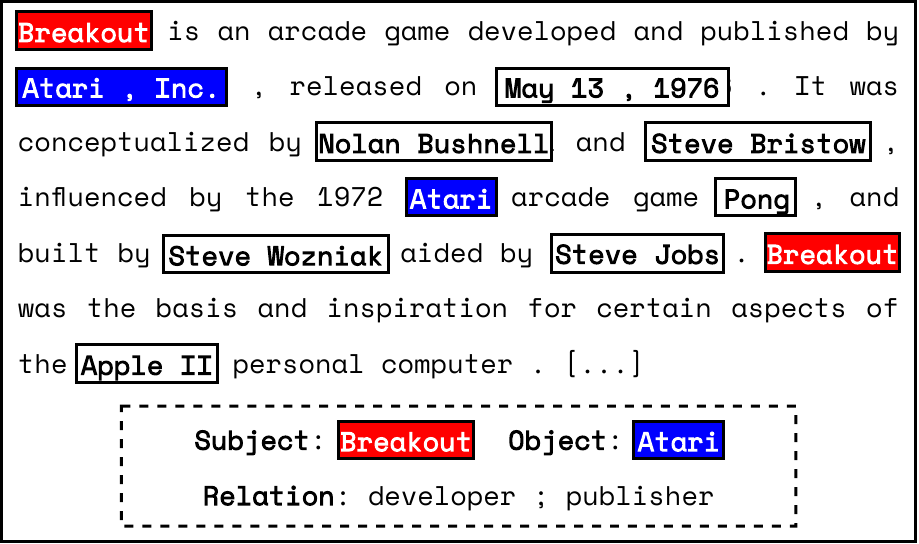}
    \vspace{-4ex}\\
    \caption{\setlength{\fboxrule}{0.75pt}\setlength{\fboxsep}{1pt}Document from the Re-DocRED \cite{tan-etal-2022-revisiting} dataset involving multiple entities and labels. Subject entity \fcolorbox{black}{red}{\color{white}{\texttt{Breakout}}} (red) and object entity \fcolorbox{black}{blue}{\color{white}{\texttt{Atari}}} (blue) express relations "developer" and "publisher". Other entities are indicated as \fcolorbox{black}{white}{\color{black}{\texttt{Mention}}} (white).}
    \vspace{-2ex}
    \label{fig:example}
\end{figure}

Document-level relation extraction is an important task in natural language processing, which involves identifying and categorising meaningful relationships between entities within a document, as exemplified in Figure \ref{fig:example}. This task is foundational to many applications, including knowledge base population and completion \cite{Banko2007OpenIE, Ji2020ASO}, information retrieval and extraction \cite{manning_raghavan_schütze_2008, theodoropoulos-etal-2021-imposing}, question answering \cite{chen-etal-2017-reading, feng-etal-2022-multi} and sentiment analysis \cite{Pang2008OpinionMA}, to name a few. 

Standard methods that approach this challenge generally employ pretrained text-based encoders \cite{devlin-etal-2019-bert, beltagy-etal-2019-scibert, zhuang-etal-2021-robustly, 10.1109/TASLP.2021.3124365}, which are responsible for capturing the nuances of information contained in the entity mentions and their contextual surroundings. Previous successful methods often then use hand-coded pooling heuristics to aggregate the information learned by the encoder, with some aimed at creating entity representations, while others directly exploiting the pattern of attention weights to capture context aware relations between entity mentions \cite{zhou2021atlop, xiao-etal-2022-sais, tan-etal-2022-document, ma-etal-2023-dreeam}. These pooling heuristics can be very effective at leveraging the information in a pretrained encoder. However, as shown in \citet{conneau-etal-2017-supervised, jia-etal-2019-document, reimers-gurevych-2019-sentence, Choi2021EvaluationOB}, the selection of an appropriate pooling function can be model-dependent, task-specific, resource-intensive and time-consuming to determine, thereby limiting flexibility.

In this paper, we address these issues with a new approach where we leverage the intrinsic graph processing capabilities of the Transformer model \cite{Vaswani2017AttentionIA}, leveraging insights from the work of \citet{mohammadshahi-henderson-2020-graph, henderson-2020-unstoppable, mohammadshahi-henderson-2021-recursive, henderson-etal-2023-transformers}. They argue that attention weights and graph relations are functionally equivalent and show how to incorporate structural dependencies between input elements by simply adding relation features to the attention functions.  Transformers easily learn to integrate these relation features into their pretrained attention functions, resulting in very successful graph-conditioned models \cite{mohammadshahi-henderson-2021-recursive,miculicich-henderson-2022-graph,mohammadshahi-henderson-2023-syntax}.
Given this effective method for integrating explicit graphs with pretrained attention functions, we propose to use the attention function itself for aggregation.  We replace the rigid pooling methods with new tokens which act as aggregation nodes, plus explicit graph relations which steer the aggregation.  

We introduce a joint text-graph Transformer model and a \underline{g}raph-\underline{a}ssisted \underline{de}clarative \underline{po}oling (GADePo) method\footnote{\url{https://github.com/idiap/gadepo}} that leverages these special tokens and graph relations, to provide an explicit high-level declarative specification for the information aggregation process. By integrating these graphs in the attention functions of a pretrained model, GADePo exploits the pretrained embeddings and attention patterns but still has the flexibility of being trained on data. This enables the pooling to be guided by domain-specific knowledge or desired outcomes but still learned by the Transformer, opening up a more customisable but still data-driven relation extraction process.

We evaluate our method across diverse datasets and models commonly employed in document-level relation extraction tasks, and show that our approach yields promising results that are consistently better than those achieved by the hand-coded pooling functions.

\paragraph{Contributions} We propose a new method for exploiting pretrained Transformer models which replaces hand-coded aggregation functions with explicit graph relations and aggregation nodes. We introduce a novel form of joint text-graph Transformer model. We evaluate our approach across various datasets and models, showing that it yields promising results that are consistently better than those achieved by hand-coded pooling functions.

\section{Related Work}
\label{sec:related_work}

\begin{figure*}
  \centering
  \includegraphics[width=\textwidth]{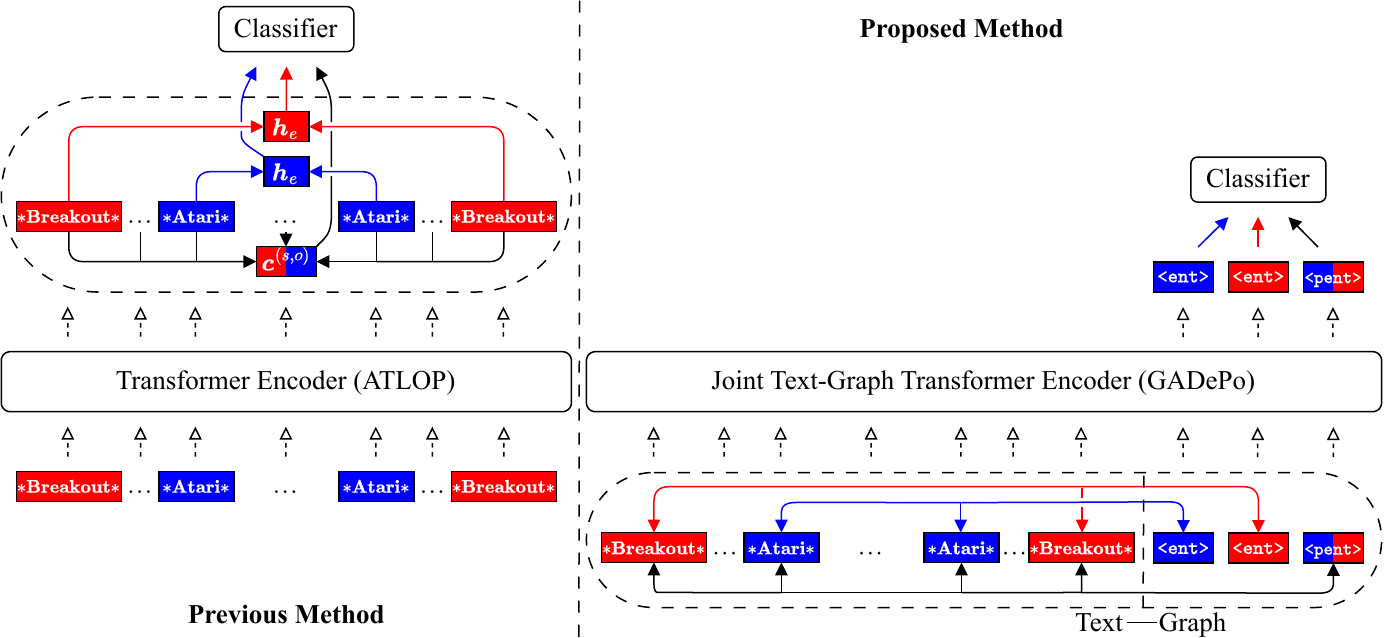}
  \vspace{-4ex}\\
  \caption{Comparison between the previous method ATLOP (left) and the proposed method GADePo (right), illustrating the document in Figure \ref{fig:example} containing two entities (red and blue), each with two mentions. In ATLOP, the mentions' encoder outputs are aggregated into entity representations $\bm{h}_e$, and the encoder's attention weights are used to identify which outputs to aggregate for entity-pair representations $\bm{c}^{(s,o)}$. In GADePo, the textual input is extended to include the graph special tokens \texttt{<ent>} for entity representations and \texttt{<pent>} for entity-pair representations, and explicit directional graph relations specify their associated mentions. A joint text-graph Transformer model is then used to encode this declarative pooling specification graph and compute the relevant aggregations.}
  \label{fig:atlop_gadepo}
\end{figure*}

In recent studies, the scope of relation extraction has been expanded to include not only individual sentences but entire documents. This extension, known as document-level relation extraction, presents a more realistic and challenging scenario as it seeks to extract relations both within sentences and across multiple sentences \cite{yao-etal-2019-docred}. 
Transformer-based \cite{Vaswani2017AttentionIA} models have shown great potential in addressing this task. 

\citet{Wang2019FinetuneBF} and \citet{tang2020hin} show that the BiLSTM-based \cite{Hochreiter1997LongSM} baselines lack the capacity to model complex interactions between multiple entities. They propose a more robust approach, which consists of using the pretrained BERT \cite{devlin-etal-2019-bert} model and a two-step prediction process, i.e., first identifying if a link between two entities exists, followed by predicting the specific relation type.

GAIN \cite{zeng-etal-2020-double} leverages BERT as a text encoder and GCNs \cite{kipf2017semi} to process two types of graphs, one at  mention level and another at entity level, showing notable performance in inter-sentence and inferential scenarios.

\citet{mohammadshahi-henderson-2020-graph, mohammadshahi-henderson-2021-recursive} propose the G2GT model and show how to leverage the intrinsic graph processing capabilities of the Transformer model by incorporating structural dependencies between input elements as features input to the self-attention weight computations.

SSAN \cite{Xu_Wang_Lyu_Zhu_Mao_2021} leverages this idea and considers the structure of entities. It employs a transformation module that creates attentive biases from this structure to regulate the attention flow during the encoding phase.

DocuNet \cite{Zhang2021DocumentlevelRE} reformulates the task as a semantic segmentation problem. It employs a U-shaped segmentation module and an encoder module to capture global interdependencies and contextual information of entities, respectively.

PL-Marker \cite{ye-etal-2022-packed} introduces a method that takes into account the interplay between spans via a neighbourhood-oriented and subject-oriented packing approach, highlighting the importance of capturing the interrelation among span pairs in relation extraction tasks.

SAIS \cite{xiao-etal-2022-sais} explicitly models key information sources such as relevant contexts and entity types. It improves extraction quality and interpretability, while also boosting performance through evidence-based data augmentation and ensemble inference.

KD-DocRE \cite{tan-etal-2022-document} proposes a semi-supervised framework with three key components. Firstly, an axial attention module enhances performance in handling two-hop relations by capturing the interdependence of entity pairs. Secondly, an adaptive focal loss solution addresses the class imbalance issue. Lastly, the framework employs knowledge distillation to improve robustness and overall effectiveness by bridging the gap between human-annotated and distantly supervised data.

DREEAM \cite{ma-etal-2023-dreeam} is a method designed to enhance document-level relation extraction by addressing memory efficiency and annotation limitations in evidence retrieval. It employs evidence as a supervisory signal to guide attention and introduces a self-training strategy to learn evidence retrieval without requiring evidence annotations.

SAIS \cite{xiao-etal-2022-sais}, KD-DocRE \cite{tan-etal-2022-document}, and DREEAM \cite{ma-etal-2023-dreeam} have been built upon the foundations of ATLOP \cite{zhou2021atlop}. ATLOP introduces two innovative techniques, adaptive thresholding, and localised context pooling, to address challenges in multi-label and multi-entity problems. Adaptive thresholding employs a learnable entities-dependent threshold, replacing the global threshold used in previous approaches for multi-label classification \cite{peng-etal-2017-cross, christopoulou-etal-2019-connecting, nan-etal-2020-reasoning, wang-etal-2020-global}. Localised context pooling leverages the attention patterns of a pretrained language model to identify and extract relevant context crucial for determining the relation between entities, using specific hand-coded pooling functions.

\section{Background}
\label{sec:background}

The foundational work of ATLOP \cite{zhou2021atlop} has been the basis of many State-of-the-Art (SotA) models \cite{xiao-etal-2022-sais,tan-etal-2022-document,ma-etal-2023-dreeam}.  
Given the problems with hand-coded pooling functions, discussed in Section~\ref{sec:introduction}, we aim to provide a new baseline that can serve as the foundation for future SotA models.
For this reason, we evaluate our proposed models by comparing them to this established baseline.
Our goal is to demonstrate that our method not only achieves results comparable to or better than ATLOP, but also offers a novel approach which addresses its limitations.  To provide a better understanding of ATLOP and its components, we present a detailed breakdown in the left portion of Figure \ref{fig:atlop_gadepo}, which we elaborate on in this section.

\subsection{Problem Formulation}
\label{subsec:problem_formulation}
The document-level relation extraction task involves analysing a document $D$ that contains a set of entities $\mathcal{E}_D {=} \{e_i\}_{i=1}^{\vert \mathcal{E}_D \vert}$. The main objective is to determine the presence or absence of various relation types between all entity pairs $(e_s, e_o)_{s,o \in \mathcal{E}_D,s \neq o}$, where the subject and object entities are denoted as $e_s$ and $e_o$, respectively. A key aspect to consider is that an entity can appear multiple times in the document, resulting in a cluster of multiple mentions $\mathcal{M}_e {=} \{m_i\}_{i=1}^{\vert \mathcal{M}_e \vert}$ for each entity $e$. The set of relations is defined as $\mathcal{R} \cup {\emptyset}$, where $\emptyset$ represents the absence of a relation, often referred to as "no-relation". Given the clusters of mentions $\mathcal{M}_{e_s}$ and $\mathcal{M}_{e_o}$, the task consists of a multi-label classification problem where there can be multiple relations between entities ${e_s}$ and ${e_o}$.

\subsection{Previous Method: ATLOP}
\label{subsec:previous_method}

\paragraph{Text Encoding} A special token $*$ is added at the start and end of every mention. Tokens $\mathcal{T}_D {=} \{t_i\}_{i=1}^{\vert \mathcal{T}_D \vert}$ are encoded via a Pretrained Language Model (PLM) as follows: 
\begin{equation}
\label{eq:plm_encoding}
    \bm{H}, \bm{A} = PLM(\mathcal{T}_D),
\end{equation}
where $\bm{H} \in \mathbb{R}^{\vert \mathcal{T}_D \vert \times d}$ and $\bm{A} \in \mathbb{R}^{\vert \mathcal{T}_D \vert \times \vert \mathcal{T}_D \vert}$ represent the token embeddings and the average attention weights of all attention heads, respectively, extracted from the last layer of the PLM.

\paragraph{Entity Embedding (EE)}
For each individual entity $e$ with mentions $\mathcal{M}_e {=} \{m_i\}_{i=1}^{\vert \mathcal{M}_e \vert}$, an entity embedding $\bm{h}_e \in \mathbb{R}^{d}$ is computed as follows:
\begin{equation}
\label{eq:logsumexp}
    \bm{h}_e = log \sum_{i=1}^{\vert \mathcal{M}_e \vert} exp(\bm{H}_{m_i}),
\end{equation}
where $\bm{H}_{m_i} \in \mathbb{R}^d$ is the embedding of the special token $*$ at the starting position of mention $m_i$. 
The choice of the $logsumexp$ pooling function is based on the research conducted by \citet{jia-etal-2019-document}. Their study offers empirical evidence that supports the use of this pooling function over others, as it facilitates accumulating weak signals from individual mentions, thanks to its smoother characteristics.

\paragraph{Localised Context Embedding (LCE)} 
ATLOP introduces the concept of localised context embedding to accommodate the variations in relevant mentions and context for different entity pairs $(e_s, e_o)$. Since the attention mechanism in the PLM captures the importance of each token within the context, it can be used to determine the context relevant for both entities. The importance of each token can be computed from the cross-token dependencies matrix $\bm{A}$ obtained in Equation \ref{eq:plm_encoding}. When evaluating entity $e_s$, the importance of individual tokens is determined by examining the cross-token dependencies across all mentions associated with $e_s$, denoted as $\mathcal{M}_{e_s}$. Initially, ATLOP collects and averages the attention $\bm{A}_{m_i} \in \mathbb{R}^{\vert \mathcal{T}_D \vert}$ at the special token $*$ preceding each mention $m_i \in \mathcal{M}_{e_s}$. This process results in $\bm{a}_s \in \mathbb{R}^{\vert \mathcal{T}_D \vert}$, which represents the importance of each token concerning entity $e_s$ (and analogously $\bm{a}_o$ for $e_o$). Subsequently, the importance of each token for a given entity pair $(e_s, e_o)$, denoted as $\bm{q}^{(s,o)} \in \mathbb{R}^{\vert \mathcal{T}_D \vert}$, is computed using $\bm{a}_s$ and $\bm{a}_o$ as follows:
\begin{equation}
    \bm{q}^{(s,o)} = \frac{\bm{a}_s \circ \bm{a}_o}{\bm{a}_s^\top \bm{a}_o},
\end{equation}
where $\circ$ represents the Hadamard product. Consequently, $\bm{q}^{(s,o)}$ represents a distribution that indicates the importance of each token for both tokens in $(e_s, e_o)$. Finally, the localised context embedding is computed as follows:
\begin{equation}
\label{eq:cso}
    \bm{c}^{(s,o)} = \bm{H}^\top\bm{q}^{(s,o)},
\end{equation}
So $\bm{c}^{(s,o)} \in \mathbb{R}^d$ corresponds to a weighted average over all token embeddings that are important for both $e_s$ and $e_o$.

\paragraph{Relation Classification and Loss Function} 
The representations $\bm{h}_{e_s}$, $\bm{h}_{e_o}$ and $\bm{c}^{(s,o)}$ are input to a relation classifier, and the full model is fine-tuned to predict the relation labels for $(e_s,e_o)$.  The relation classifier and its loss function are detailed in Appendix Subsection \ref{subsec:atlop_relation_classification_and_loss_function}.

\section{Proposed Method: GADePo}
\label{sec:proposed_method}

We propose to avoid the reliance on the EE (i.e., $\bm{h}_e$) and LCE (i.e., $\bm{c}^{(s,o)}$) heuristic aggregation functions by leveraging Transformers' attention functions to do aggregation.  Given the observation of \citet{henderson-2020-unstoppable, mohammadshahi-henderson-2020-graph, mohammadshahi-henderson-2021-recursive, henderson-etal-2023-transformers} that attention weights and graph relations are functionally equivalent, we introduce the inductive biases of EE and LCE directly into the model's input as graph relations.

Our proposed \underline{g}raph-\underline{a}ssisted \underline{de}clarative \underline{po}oling (GADePo) method replaces the hand-coded aggregation functions EE and LCE with a declarative graph specification. By using the intrinsic graph processing capabilities of the Transformer model, the specified graph serves as an explicit high-level directive for the information aggregation process of the Transformer.  By inputting the graph relations to the Transformer's self-attention layers, GADePo enables the aggregation to be steered by domain-specific knowledge or desired outcomes, while still allowing it to be learned by the Transformer, opening up the possibility for a more tailored and customised yet data-driven relation extraction.

\noindent Our GADePo model is illustrated in the right portion of Figure \ref{fig:atlop_gadepo}. We address both EE and LCE with the introduction of two special tokens, \texttt{<ent>} (i.e., entity) and \texttt{<pent>} (i.e., pair entity), and two explicit graph relations of types $\texttt{<ent>} \longleftrightarrow *$ and $\texttt{<pent>} \longleftrightarrow *$ in both directions, where $*$ represents the special token at the starting position of a specific mention. The set of relations is specified as $\bm{c}_{ij} \in \mathcal{C}$ which each identify the relation label from $i$ to $j$. Each of these relation labels is associated with an embedding vector of dimension $d$, as are the special token inputs \texttt{<ent>} and \texttt{<pent>}. These two special tokens are added to the PLM's vocabulary of input tokens, while relation label embeddings are input to the self-attention functions for every pair of related tokens.  These new embeddings represent learnable parameters that are trained during the PLM fine-tuning on the downstream tasks. As reported in Appendix Subsection \ref{subsec:gadepos_extra_parameters}, GADePo adds a negligible number of extra parameters, namely only the special token inputs and the graph directional relation inputs. 

\paragraph{Special Token \texttt{<ent>}} 
To tackle the EE pooling function, we add to the input tokens $\mathcal{T}_D$ as many \texttt{<ent>} special tokens as entities in the document. This way each entity $e$ has a corresponding entity token \texttt{<ent>} in the input. We connect each \texttt{<ent>} token with its corresponding cluster of mentions $\mathcal{M}_e {=} \{m_i\}_{i=1}^{\vert \mathcal{M}_e \vert}$, and vice-versa. The two graph relations we use are thus $\texttt{<ent>} \longrightarrow *$ and $* \longrightarrow \texttt{<ent>}$, where $*$ represents the special token at the starting position of mention $m_i$.  Each \texttt{<ent>} token receives the same \texttt{<ent>} embedding, with no positional encoding, since each one collectively represents a set of mentions from different positions in the input graph.  These identical inputs are only disambiguated through the connections to and from mentions expressed as the $\texttt{<ent>} \longrightarrow *$ and $* \longrightarrow \texttt{<ent>}$ graph relations.  These relations tell the self-attention mechanism to use the \texttt{<ent>} token to aggregate information from the associated mentions, and thus the \texttt{<ent>} tokens have a direct correspondence to the computed $\bm{h}_e$ in Equation \ref{eq:logsumexp}.

\paragraph{Special Token \texttt{<pent>}} 
ATLOP performs information filtering by calculating via Equation \ref{eq:cso} a localised context embedding (LCE) $\bm{c}^{(s,o)}$ that is dependent on the cross-token attention matrix $\bm{A}$ output by the PLM. The intuition behind it is that the dependencies between different tokens are encoded as attention weights. We propose a straightforward adjustment of the input graph used for the EE pooling to effectively model and capture these dependencies. To address the LCE pooling function, we add to the input tokens $\mathcal{T}_D$ as many \texttt{<pent>} special tokens as the number of all possible pairs of entities. Each special token \texttt{<pent>} thus refers to a pair of entities $(e_s, e_o)$. We connect each \texttt{<pent>} token with each mention in the two clusters of mentions $\mathcal{M}_{e_s} {=} \{m_i\}_{i=1}^{\vert \mathcal{M}_{e_s} \vert}$ and $\mathcal{M}_{e_o} {=} \{m_i\}_{i=1}^{\vert \mathcal{M}_{e_o} \vert}$ and vice-versa. Since the attention weights used in LCE are computed from these mention embeddings, we expect that they are sufficient for the Transformer to learn to find the relevant contexts. The two graph relations we use are thus $\texttt{<pent>} \longrightarrow *$ and $* \longrightarrow \texttt{<pent>}$.  Analogously to the \texttt{<ent>} tokens, the \texttt{<pent>} tokens all receive the same \texttt{<pent>} embedding, with no positional embeddings, and thus are only disambiguated by their different $\texttt{<pent>} \longrightarrow *$ and $* \longrightarrow \texttt{<pent>}$ graph relations.  These relations tell the \texttt{<pent>} token to pay attention to its associated mentions, which in turn allows it to find the relevant context shared by these mentions.  Thus, each \texttt{<pent>} token can be seen as having a direct correspondence to the computed $\bm{c}^{(s,o)}$ in Equation~\ref{eq:cso}. 

All equations relative to the relation classification and the corresponding loss function reported in Appendix Subsection \ref{subsec:atlop_relation_classification_and_loss_function} remain valid as we merely substitute the hand-coded computations of $\bm{h}_e$ and $\bm{c}^{(s,o)}$ with the embeddings of \texttt{<ent>} and \texttt{<pent>}, respectively.  

\begin{table*}
\centering
\begin{tabular}{l|l|cc|ccc}
\multicolumn{2}{c}{} & \multicolumn{2}{c}{\textbf{Re-DocRED}} & \multicolumn{3}{c}{\textbf{HacRED}} \\
\textbf{Model} & \textbf{Aggregation} & Ign $F_1$ & $F_1$ & $P$ & $R$ & $F_1$ \\
\hline
ATLOP$^{\star}$ & $\bm{h}_e$ & $75.27$ & $75.92$ & $\textbf{76.27}$ & $76.83$ & $76.55$ \\
GADePo (ours) & \texttt{<ent>} & $\textbf{75.55}$ & $\textbf{76.38}$ & $74.13$ & $\textbf{79.46}$ & $\textbf{76.70}$ \\
\hline
ATLOP$^{\bullet,\diamond}$ & $\bm{h}_e$ ; $\bm{c}^{(s,o)}$ & $76.82$ & $77.56$ & $77.89$ & $76.55$ & $77.21$ \\
ATLOP$^{\star}$ & $\bm{h}_e$ ; $\bm{c}^{(s,o)}$ & $77.62$ & $78.38$ & $76.36$ & $78.86$ & $77.59$ \\
GADePo (ours) & \texttt{<ent>} ; \texttt{<pent>} & $\textbf{77.70}$ & $\textbf{78.40}$ & $\textbf{78.27}$ & $\textbf{79.03}$ & $\textbf{78.65}$ 
\vspace{-1ex}
\end{tabular} 
\caption{Comparative analysis between the previous method ATLOP and the proposed method GADePo on the test set. ATLOP$^{\star}$ indicates our reimplementation of the previous method. For Re-DocRED and HacRED we report in percentage the results obtained by \citet{tan-etal-2022-revisiting} (ATLOP$^{\bullet}$) and \citet{cheng-etal-2021-hacred} (ATLOP$^{\diamond}$), respectively. The results are reported in terms of $F_1$ scores, Precision ($P$), and Recall ($R$), following the same metrics reported in prior research specific to each dataset. Ign $F_1$ denotes the $F_1$ score that excludes relational facts shared between the training and evaluation sets. We also comply with the standard practice where test scores are determined based on the best checkpoint from five training runs with distinct random seeds.}
\label{tab:redocred_hacred_results}
\end{table*}

\paragraph{Text-Graph Encoding} 
We follow \citet{mohammadshahi-henderson-2020-graph, mohammadshahi-henderson-2021-recursive, henderson-etal-2023-transformers} in leveraging the intrinsic graph processing capabilities of the Transformer model by incorporating graph relations as relation embeddings input to the self-attention function.  For every pair of input tokens $ij$, the pre-softmax attention weight $e_{ij}\in \mathbb{R}$ is computed from both the respective token embeddings $\bm{x}_i, \bm{x}_j \in \mathbb{R}^d$, and an embeddings of the graph relation $\bm{c}_{ij}$ between the $i$-th and $j$-th tokens.
However, we change the attention weight computation to:
\begin{equation}
\label{eq:attention_biasing}
    e_{ij} = \frac{ \bm{x}_i\bm{W}_Q ~ \diag(\LN(\bm{c}_{ij}\bm{W}_C)) ~ (\bm{x}_j\bm{W}_K)^\top }{\sqrt{d}},
\end{equation}
where 
$\bm{W}_Q, \bm{W}_K \in \mathbb{R}^{d \times d}$ represent the query and key matrices, respectively.  $\bm{c}_{ij} \in \{0, 1\}^{\vert \mathcal{C} \vert}$ represents a $0/1$ encoded label of the graph relation between the $i$-th and $j$-th input elements, and $\bm{W}_C \in \mathbb{R}^{\vert \mathcal{C} \vert \times d}$ represents the relations' embedding matrix, so $\bm{c}_{ij}\bm{W}_C$ is the embedding of the relation between $i$ and $j$. Finally, $\LN$ stands for the $LayerNorm$ operation and $\diag$ returns a diagonal matrix.  

Compared to the standard attention function, where $e_{ij} = \bm{x}_i\bm{W}_Q (\bm{x}_j\bm{W}_K)^\top / \sqrt{d}$, the relation embedding determines a weighting of the different dimensions. This is a novel way to condition on the relation embedding compared to the original formulation, which only models query-relation interactions \cite{mohammadshahi-henderson-2020-graph}.  This change is motivated by our task requiring a more flexible formulation which models query-relation-key interactions via a multiplicative mechanism, without requiring a full ${d \times d}$ matrix of bi-linear parameters. This way, a key will be relevant to a query only when both agree on the relation.  {In preliminary experiments, we explored various methods for biasing attention and found that the formulation presented in Equation \ref{eq:attention_biasing} produced the best results.}

\section{Experiments}
\label{sec:experiments}

\begin{figure*}
  \centering
  \includegraphics[width=\textwidth]{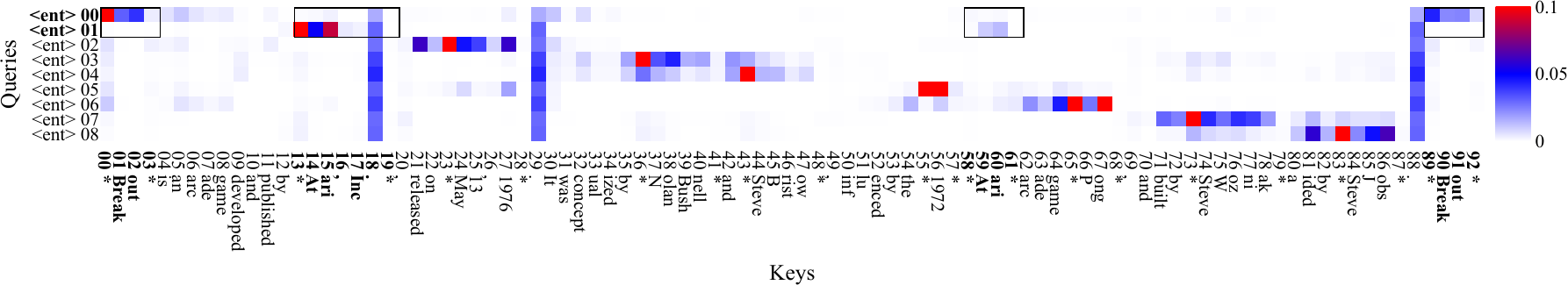}
  \vspace{-4ex}\\
  \caption{Attention weights $\bm{A}$ from GADePo via Equation \ref{eq:plm_encoding} for the document in Figure \ref{fig:example}. For clarity, only  a subset of \texttt{<ent>} and document tokens are shown on the $y$-axis (queries) and $x$-axis (keys), respectively.}
  \label{fig:heatmap}
\end{figure*}

\subsection{Datasets and Models}
\label{subsec:datasets_and_models}

\paragraph{Re-DocRED} \cite{tan-etal-2022-revisiting} is a revisited version of the DocRED \cite{yao-etal-2019-docred} dataset. It is built from English Wikipedia and Wikidata and contains both distantly-supervised and human-annotated documents with named entities, coreference data, and intra- and inter-sentence relations, supported by evidence. It requires analysing multiple sentences to identify entities, establish their relationships, and integrate information from the entire document. We comply with the model used by the authors and employ the RoBERTa\textsubscript{\textsc{large}} \cite{zhuang-etal-2021-robustly} model in our experiments.

\paragraph{HacRED} \cite{cheng-etal-2021-hacred} is a large-scale, high-quality Chinese document-level relation extraction dataset, with a special focus on practical hard cases. As the authors did not provide specific information about the model used in their study, we conducted our experiments using the Chinese BERT\textsubscript{\textsc{base}} with whole word masking model \cite{10.1109/TASLP.2021.3124365}.

\paragraph{Datasets statistics} Re-DocRED and HacRED exhibit notable distinctions in their statistics, as summarised in Table \ref{tab:redocred_hacred_datasets_statistics}. Re-DocRED comprises a larger number of facts, entities per document, and relations compared to HacRED. This indicates a potentially richer and more extensive dataset in terms of factual information and relationship types. However, HacRED contains more documents and may present a broader range of scenarios for relation extraction, including more challenging cases, as it has been specifically created with a focus on practical hard cases.

\begin{table}[H]
    \centering
    \begin{tabular}{l|c|c}
    \textbf{Statistic} & \textbf{Re-DocRED} & \textbf{HacRED} \\
    \hline
    Facts & 120,664 & 65,225 \\
    Relations & 96 & 26 \\
    Documents & 4,053 & 9,231 \\
    Average Entities & 19.4 & 10.8 
    \vspace{-1ex}
    \end{tabular}
    \caption{Re-DocRED and HacRED human-annotated datasets statistics.}
\label{tab:redocred_hacred_datasets_statistics}
\end{table}

\subsection{Results and Discussion}
\label{subsec:results_and_discussion}

\begin{figure*}
    \centering
    \begin{subfigure}{0.49\textwidth}
    \centering
    \includegraphics[width=\linewidth]{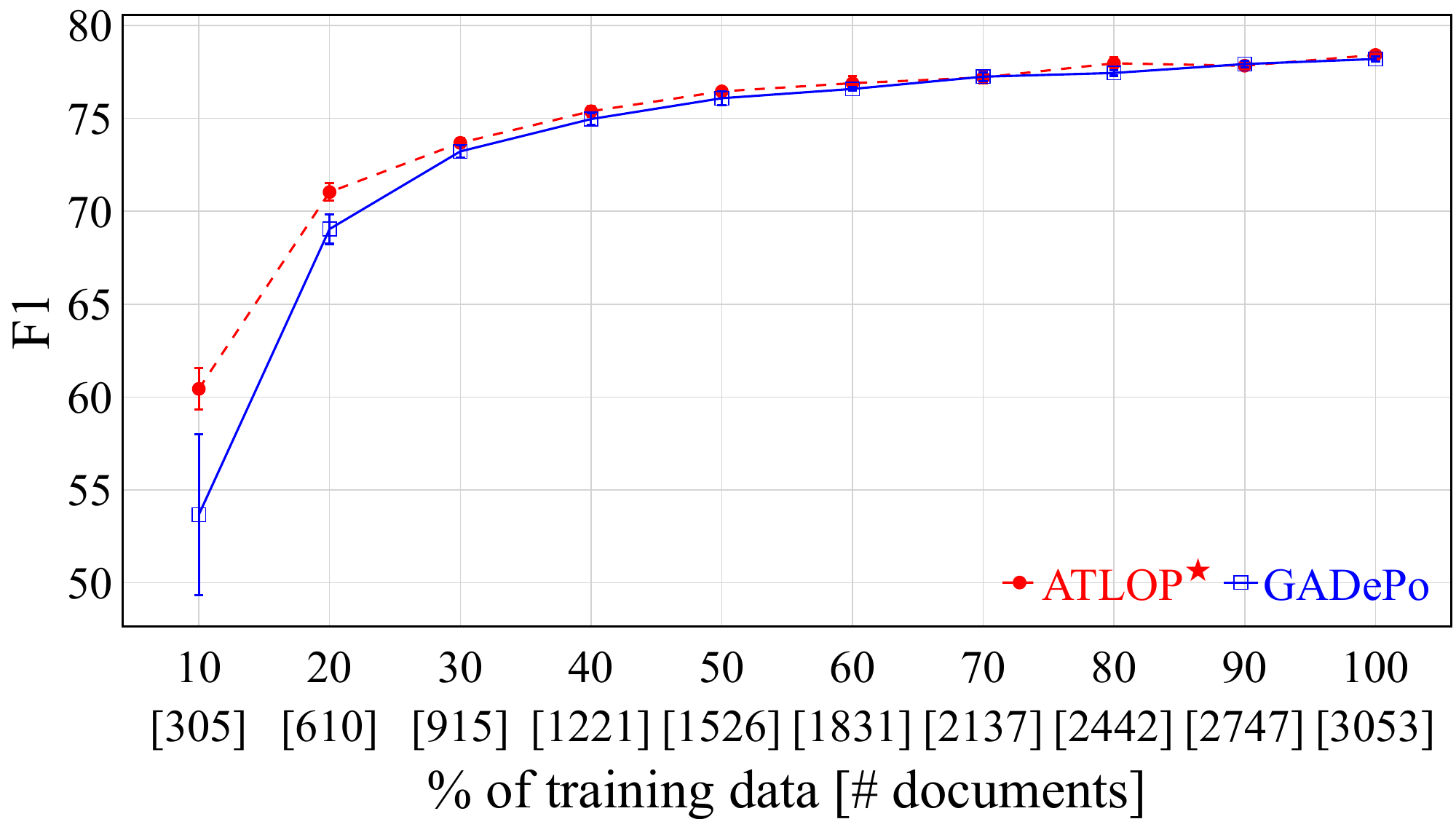}
    \caption{Re-DocRED}
    \label{subfig:redocred_f1}
    \end{subfigure}
    \hfill
    \begin{subfigure}{0.49\textwidth}
    \centering
    \includegraphics[width=\linewidth]{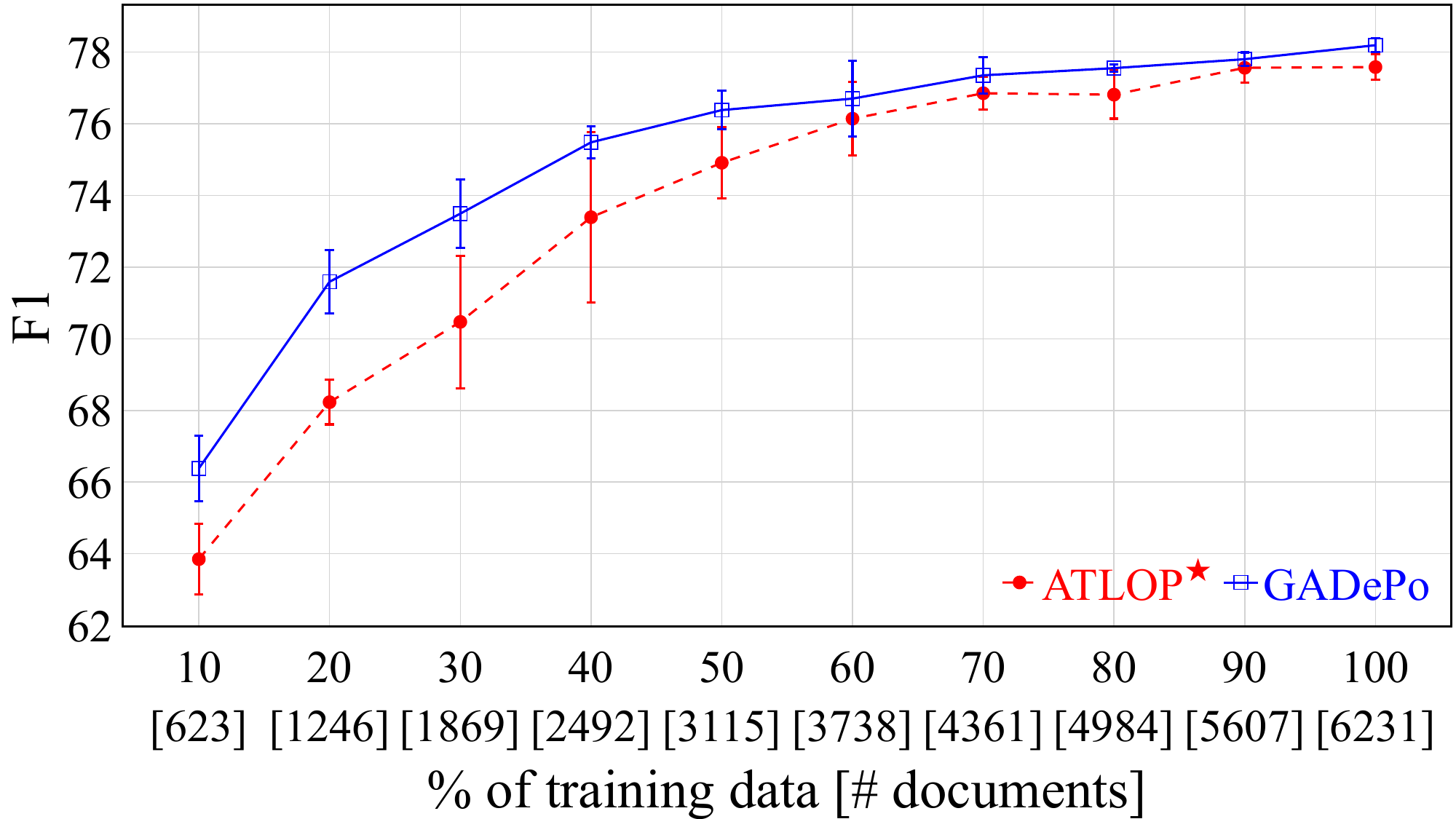}
    \caption{HacRED}
    \label{subfig:hacred_f1}
    \end{subfigure}
    \vspace{-1ex}\\
    \caption{Performance of ATLOP$^{\star}$ ($\bm{h}_e$ ; $\bm{c}^{(s,o)}$) and GADePo (\texttt{<ent>} ; \texttt{<pent>}) on the development set under varying data availability conditions on Re-DocRED (\ref{subfig:redocred_f1}) and HacRED (\ref{subfig:hacred_f1}). The $x$-axis represents the percentage and number of documents from the training dataset, while the $y$-axis displays the $F_1$ score in percentage. Each point on the graph represents the mean value, while error bars indicate the standard deviation derived from five distinct training runs with separate random seeds.}
    \label{fig:redocred_hacred_data_ablation_study}
\end{figure*}

We follow the standard practice from prior research and report the results of our experiments on the Re-DocRED and HacRED datasets in Table \ref{tab:redocred_hacred_results} and Figure \ref{fig:redocred_hacred_data_ablation_study}. For all datasets and models, we provide our reimplementation of the ATLOP baseline (indicated as ATLOP$^{\star}$), which achieves or surpasses previously reported results for ATLOP, and compare the proposed GADePo model against this model. We evaluate all datasets using the $F_1$ metric. For Re-DocRED, Ign $F_1$ (or Ignored $F_1$) is also reported, and refers to the $F_1$ score that excludes relational facts that are shared between the training and development/test sets. This is done to avoid potential biases in the evaluation metrics due to overlap in content between the sets, which might not reflect the model's ability to generalise to truly unseen data. For HacRED, we adhere to the format introduced by \citet{cheng-etal-2021-hacred} and report also the Precision ($P$) and Recall ($R$) metrics. We comply with previous research and report the test score achieved by the best checkpoint on the development set. In Appendix Subsection \ref{subsec:additional_results}, we additionally present the mean and standard deviation on the development set, calculated from five training runs with distinct random seeds. We also provide in Appendix Subsection \ref{subsec:additional_results}, the same set of experiments conducted on the original DocRED dataset. Training details and hyperparameters are outlined in Appendix Subsection \ref{subsec:training_details}.

\paragraph{Re-DocRED Results} We evaluate our proposed GADePo method against the previous ATLOP method in two stages, first comparing the use of \texttt{<ent>} tokens against the use of EE pooling ($\bm{h}_e$), and then comparing our full model against the full ATLOP model, including \texttt{<pent>} tokens and LCE pooling ($\bm{c}^{(s,o)}$), respectively.

Table \ref{tab:redocred_hacred_results} highlights the effectiveness of our proposed method. When comparing $\bm{h}_e$ with \texttt{<ent>}, we observe a noticeable improvement in both Ign $F_1$ and $F_1$ scores, achieving $75.55\%$ and $76.38\%$ respectively, compared to $75.27\%$ and $75.92\%$ attained by ATLOP$^{\star}$. This demonstrates the practical utility of employing the special token \texttt{<ent>} for information aggregation. This is illustrated in the attention weights heatmap in Figure \ref{fig:heatmap}. Incorporating $\bm{c}^{(s,o)}$ and \texttt{<pent>} into the comparison, GADePo maintains performance parity with the significantly enhanced ATLOP$^{\star}$, which outperformed ATLOP$^{\bullet}$ from \citet{tan-etal-2022-revisiting}. The latter improvement suggests that a more refined hyperparameter search can lead to performance gains, as evidenced by the increase in $F_1$ score from $77.56\%$ to $78.38\%$. GADePo achieves an $F_1$ score of $78.40\%$, affirming its competitive edge and the effectiveness of employing \texttt{<pent>} for aggregation.

\begin{table}[!h]
    \centering
    \scalebox{0.9}{
        \begin{tabular}{l|l|cc}
        \textbf{Model} & \textbf{Aggregation} & Ign $F_1$ & $F_1$ \\
        \hline
        ATLOP$^{\star}$ & $\bm{h}_e$ & $76.39$ & $76.97$ \\
        GADePo (ours) & \texttt{<ent>} & $\textbf{76.99}$ & $\textbf{77.79}$ \\
        \hline
        ATLOP$^{\star}$ & $\bm{h}_e$ ; $\bm{c}^{(s,o)}$ & $77.49$ & $78.09$ \\
        GADePo (ours) & \texttt{<ent>} ; \texttt{<pent>} & $\textbf{77.50}$ & $\textbf{78.15}$ 
        \vspace{-1ex}
        \end{tabular}
    }
    \caption{Re-DocRED results on the test set following prior finetuning on the distantly supervised dataset.}
\label{tab:redocred_distant_supervision_results}
\end{table}

\noindent Table \ref{tab:redocred_distant_supervision_results} illustrates the results obtained with prior finetuning on the distantly supervised dataset, which contains approximately $100K$ documents \cite{yao-etal-2019-docred}. Interestingly, distant supervision appears to have a slightly negative impact on the results of both methods when incorporating $\bm{c}^{(s,o)}$ or \texttt{<pent>}. However, it proves to be highly beneficial when utilising solely $\bm{h}_e$ or \texttt{<ent>} for aggregation. This suggests that although distant supervision might introduce noise into the training process, it can also provide valuable information that improves model generalisation, particularly when leveraging simpler feature representations like $\bm{h}_e$ and \texttt{<ent>}, possibly due to their robustness in capturing essential information amidst noise.

\paragraph{HacRED Results} We observe a similar pattern to Re-DocRED, with ATLOP$^{\star}$ displaying a slight performance advantage over ATLOP$^{\diamond}$ from \citet{cheng-etal-2021-hacred} (Table~\ref{tab:redocred_hacred_results}). On this dataset, GADePo shows a significantly improved performance, primarily driven by a substantial increase in Recall ($R$), indicating that the GADePo model is more effective at identifying relevant instances. As already reported for the Re-DocRED dataset, the performance boost after the inclusion of $\bm{c}^{(s,o)}$ and \texttt{<pent>} into ATLOP$^{\star}$ and GADePo, respectively, highlight the significant contributions of these features. GADePo outperforms ATLOP$^{\star}$ with an $F_1$ score of $78.65\%$ compared to $77.59\%$. This larger improvement on HacRED suggests that GADePo is better at handling challenging cases, which is not surprising given its greater flexibility over the fixed pooling functions of ATLOP.

\paragraph{Data Ablation}

To evaluate the models' sensitivity to dataset size,
the performance evaluation depicted in Figure \ref{fig:redocred_hacred_data_ablation_study} compares ATLOP$^{\star}$ ($\bm{h}_e$ ; $\bm{c}^{(s,o)}$) and GADePo (\texttt{<ent>} ; \texttt{<pent>}) on the development set, considering different levels of training data availability on the Re-DocRED and HacRED datasets. Accuracies generally converge as the dataset sizes increase, but on the challenging cases of HacRED, GADePo maintains a substantial advantage across the full range.  On Re-DocRED, GADePo catches up with and slightly outperforms ATLOP$^{\star}$ as data size increases.
This lower performance on smaller datasets is presumably because GADePo must learn how to exploit the graph relations to the special tokens \texttt{<ent>} and \texttt{<pent>} and pool information through them, whereas for ATLOP this pooling is hand-coded.
On the Re-DocRED dataset, ATLOP$^{\star}$ appears to have relatively consistent variance, while GADePo exhibits higher variance in the smaller training sets, while on the HacRED dataset, GADePo is significantly more stable for smaller datasets.

The data ablation analysis shows that the performance of hand-coded pooling functions can be dataset-specific, which restricts their adaptability. In contrast, GADePo consistently outperforms its hand-coded counterparts on larger datasets, and matches them on all but some smaller datasets, presumably due to its flexibility.  This pattern suggests that GADePo has a greater potential for optimisation, particularly on larger datasets. This is supported by GADePo's better performance on HacRED, which is both larger and designed to be more challenging than Re-DocRED.

\section{Conclusion}
\label{sec:conclusion}

In this paper we proposed a novel approach to document-level relation extraction, challenging the conventional reliance on hand-coded pooling functions for information aggregation. Our method leverages the power of Transformer models by incorporating explicit graph relations as instructions for information aggregation. By combining graph processing with text-based encoding, we introduced the \underline{g}raph-\underline{a}ssisted \underline{de}clarative \underline{po}oling (GADePo) specification, which allows for more flexible and customisable specification of pooling strategies which are still learned from data.

We conducted evaluations using diverse datasets and models commonly employed in document-level relation extraction tasks. The results of our experiments demonstrated that our approach achieves promising performance that is comparable to or better than that of hand-coded pooling functions. This suggests that our method can serve as a viable basis for other relation extraction methods, providing a more adaptable and tailored approach.
In particular, recent methods have improved performance by exploiting information about evidence, which can naturally be incorporated in our graph-based approach.

\section*{Limitations}
\label{sec:limitations}

While the proposed GADePo model offers a promising and innovative approach to relation extraction, there are issues which the current study does not address. According to the data in Appendix Table \ref{tab:redocred_hacred_datasets_statistics}, the average number of entities per document across datasets is approximately $15$. This means that, on average, there will be an additional $15$ \texttt{<ent>} tokens and $105$ \texttt{<pent>} tokens. Given that the maximum allowable input length for the models is $512$ tokens, the inclusion of these extra tokens results in roughly a $3\%$ and $20\%$ increase in the overall input length for \texttt{<ent>} and \texttt{<pent>}, respectively. It's evident that the majority of the increase in input length is due to the quadratic number of \texttt{<pent>} special tokens, but we believe that an appropriate pruning strategy could easily reduce this number to linear in the number of entities without degrading accuracy. One such pruning strategy could involve an \texttt{<ent>}-only model with a binary classifier which is trained to predict pairs of related entities.  This model could then be used to prune the set of candidate entity pairs for the final relation classification, with \texttt{<pent>} tokens being instantiated only for these candidate pairs. We have chosen to leave this approach as a potential avenue for future work, opting instead to focus on demonstrating the promise of the current simpler formulation.

\section*{Ethics Statement}
\label{sec:ethics_statement}

We do not anticipate any ethical concerns related to our work, as it primarily presents an alternative approach to a previously proposed method. Our main contribution lies in introducing a novel methodology for relation extraction. In our experiments, we use the same datasets and pretrained models as previous research, all of which are publicly available. However, it is important to acknowledge that these datasets and models may still require further examination for potential fairness issues and the knowledge they encapsulate.

\section*{Acknowledgements}
\label{sec:acknowledgement}
We extend our special gratitude to the Swiss National Science Foundation (SNSF) and Research Foundation – Flanders (FWO) for funding this work under grants 200021E\_189458 and G094020N.

\bibliography{anthology,custom}

\appendix

\section{Appendix}
\label{sec:appendix}

\subsection{ATLOP: Relation Classification and Loss Function}
\label{subsec:atlop_relation_classification_and_loss_function}

\paragraph{Relation Classification} To predict the relation between the subject entity $e_s$ and object entity $e_o$, ATLOP first generates context-aware subject and object representations as follows:
\begin{equation}
\label{eq:relation_classification_subject}
    \bm{z}_s = tanh(\bm{W}_s[\bm{h}_{e_s};\bm{c}^{(s,o)}] + \bm{b}_s)
\end{equation}
\begin{equation}
\label{eq:relation_classification_object}
    \bm{z}_o = tanh(\bm{W}_o[\bm{h}_{e_o};\bm{c}^{(s,o)}] + \bm{b}_o),
\end{equation}
where $\bm{z}_s, \bm{z}_o \in \mathbb{R}^d$, $[\cdot;\cdot]$ represents the concatenation of two vectors, and $\bm{W}_s, \bm{W}_o \in \mathbb{R}^{d \times 2d}$ together with $\bm{b}_s, \bm{b}_o \in \mathbb{R}^d$ are trainable parameters. Then, the entity pair representation is computed as:
\begin{equation}
    \bm{x}^{(s,o)} = \bm{z}_s \otimes \bm{z}_o,
\end{equation}
where $\bm{x}^{(s,o)} \in \mathbb{R}^{d^2}$ and $\otimes$ stands for the vectorised Kronecker product. Finally, relation scores are computed as:
\begin{equation}
    \bm{y}^{(s,o)} = \bm{W}_r \bm{x}^{(s,o)} + \bm{b}_r,
\end{equation}
where $\bm{y}^{(s,o)} \in \mathbb{R}^{\vert \mathcal{R} \vert}$, with $\bm{W}_r \in \mathbb{R}^{\vert \mathcal{R} \vert \times d^2}$ and $\bm{b}_r \in \mathbb{R}^{\vert \mathcal{R} \vert}$ representing learnable parameters. The probability of relation $r \in \mathcal{R}$ between the subject and object entities is computed as follows:
\begin{equation}
    P(r \vert s,o) = \sigma(\bm{y}^{(s,o)}),
\end{equation}
where $\sigma$ is the sigmoid function. To reduce the number of parameters in the classifier, a grouped function is used, which splits the embedding dimensions into $k$ equal-sized groups and applies the function within the groups as follows:
\begin{equation}
\label{eq:relation_classification_subject_block}
    \bm{z}_s = [\bm{z}_s^1;\ldots;\bm{z}_s^k]
\end{equation}
\begin{equation}
\label{eq:relation_classification_object_block}
    \bm{z}_o = [\bm{z}_o^1;\ldots;\bm{z}_o^k]
\end{equation}
\begin{equation}
   \bm{x}^{(s,o)} = [\bm{x}^{(s,o)^1};\ldots;\bm{x}^{(s,o)^k}]
\end{equation}
\begin{equation}
    \bm{y}^{(s,o)} = \sum_{i=1}^k \bm{W}_r^i \bm{x}^{(s,o)^i} + \bm{b}_r,
\end{equation}
where $\bm{z}_s^i, \bm{z}_o^i \in \mathbb{R}^{d/k}$, $\bm{x}^{(s,o)^i} \in \mathbb{R}^{d^2/k}$, and $\bm{W}_r^i \in \mathbb{R}^{\vert \mathcal{R} \vert \times d^2/k}$. This way, the number of parameters can be reduced from $d^2$ to $d^2/k$.

\paragraph{Loss Function} ATLOP introduces the adaptive thresholding loss concept. This approach involves training a model to learn a hypothetical threshold class $T\!H$, which dynamically adjusts for each relation class $r \in \mathcal{R}$. During training, for each entity pair $(e_s, e_o)$, the loss enforces the model to generate scores above $T\!H$ for positive relation classes $\mathcal{R}_P$ and scores below $T\!H$ for negative relation classes $\mathcal{R}_N$. The loss is computed as follows:
\begin{equation}
\begin{aligned}
    \mathcal{L} = & - \sum_{s \neq o} \sum_{r \in \mathcal{R}_P} \frac{exp(y_r^{(s,o)})}{\sum_{r^{'} \in \mathcal{R}_P \cup \{T\!H\}} exp(y_{r'}^{(s,o)})} \\
    & - \frac{exp(y_{T\!H}^{(s,o)})}{\sum_{r^{'} \in \mathcal{R}_N \cup \{T\!H\}} exp(y_{r'}^{(s,o)})}
\end{aligned}
\end{equation}

\subsection{GADePo's Extra Parameters}
\label{subsec:gadepos_extra_parameters}

GADePo introduces few extra parameters to the PLM. The amount of parameters is reported in Table \ref{tab:gadepos_extra_parameters}.

\begin{table}[H]
\centering
\begin{tabular}{lccc}
\textbf{Parameter} & \multicolumn{2}{c}{\textbf{Model}} \\
 & RoBERTa\textsubscript{\textsc{large}} & BERT\textsubscript{\textsc{base}} \\
\hline
\texttt{<ent>} & 1024 & 768 \\
\texttt{<pent>} & 1024 & 768 \\
$\texttt{<ent>} \longrightarrow *$  & 24 $\times$ 1024 & 12 $\times$ 768 \\
$* \longrightarrow \texttt{<ent>}$  & 24 $\times$ 1024 & 12 $\times$ 768 \\
$\texttt{<pent>} \longrightarrow *$  & 24 $\times$ 1024 & 12 $\times$ 768 \\
$* \longrightarrow \texttt{<pent>}$  & 24 $\times$ 1024 & 12 $\times$ 768 \\
\hline
\textbf{Total} & 100,352 & 38,400 
\vspace{-1ex}
\end{tabular}
\caption{GADePo's extra parameters count.}
\label{tab:gadepos_extra_parameters}
\end{table}

The introduction of these parameters results in only a minimal increase in the overall parameter count of the models. Specifically, GADePo's augmentation amounts to a mere $0.036\%$ increase over the BERT\textsubscript{\textsc{base}} model. In contrast, even a slight increase of just one unit in BERT\textsubscript{\textsc{base}}'s hidden dimensions would result in a $0.139\%$ parameter increase, which is roughly four times greater than the augmentation introduced by GADePo. Given that such a small change is incompatible with other architectural constraints, such as the number of heads, it is implausible that this minimal augmentation would solely account for the observed performance gains.

This indicates that the performance improvements are largely due to the effective inductive bias introduced by GADePo, rather than the increase in parameter count. The same rationale applies to the results observed with RoBERTa\textsubscript{\textsc{large}}.

\subsection{Training Details}
\label{subsec:training_details}

We generally comply with the hyperparameters of ATLOP and set the output dimension in Equation \ref{eq:relation_classification_subject} and Equation \ref{eq:relation_classification_object} to 768. We also set the block size in Equation \ref{eq:relation_classification_subject_block} and Equation \ref{eq:relation_classification_object_block} to 64, i.e., $k=12$.

In all our experiments we perform early stopping on the development set based on the Ign $F_1 + F_1$ score for DocRED and Re-DocRED, and $F_1$ score for HacRED. The five different seeds we use are $\{73, 21, 37, 7, 3\}$.

We use RAdam \cite{Liu2020On} as our optimiser. On the RoBERTa\textsubscript{\textsc{large}} based models we train for 8 epochs and set the learning rates to $3e^{-5}$ and $1e^{-4}$ for the PLM parameters and the new additional parameters, respectively. On the BERT\textsubscript{\textsc{base}} based models we train for 10 epochs and set the learning rates to $1e^{-5}$ and $1e^{-4}$ for the PLM parameters and the new additional parameters, respectively. We use a cosine learning rate decay throughout the training process.

In all our experiments the batch size is set to 4 for ATLOP and 2 for GADePo, with gradient accumulation set to 1 and 2, for ATLOP and GADePo, respectively. We clip the gradients to a max norm of 1.0. All models are trained with mixed precision.

We run our experiments on two types of GPUs, namely the NVIDIA V100 32GB for the RoBERTa\textsubscript{\textsc{large}} based models and NVIDIA RTX 3090 24GB for the BERT\textsubscript{\textsc{base}} based models, respectively.

We use PyTorch \cite{Paszke_PyTorch_An_Imperative_2019}, Lightning \cite{Falcon_PyTorch_Lightning_2019}, and Hugging Face's Transformers \cite{wolf-etal-2020-transformers} libraries to develop our models.

\begin{table*}
\centering
\begin{tabular}{l|l|cc|cc}
\multicolumn{2}{c}{} & \multicolumn{2}{c}{\textbf{Dev}} & \multicolumn{2}{c}{\textbf{Test}} \\
\textbf{Model} & \textbf{Aggregation} & Ign $F_1$ & $F_1$ & Ign $F_1$ & $F_1$ \\
\hline
ATLOP$^{\star}$ & $\bm{h}_e$ & $75.46 \pm 0.16$ & $76.16 \pm 0.16$ & $75.27$ & $75.92$ \\
GADePo (ours) & \texttt{<ent>} & $75.46 \pm 0.20$ & $76.31 \pm 0.24$ & $\textbf{75.55}$ & $\textbf{76.38}$ \\
\hline
ATLOP$^{\bullet}$ & $\bm{h}_e$ ; $\bm{c}^{(s,o)}$ & $76.79$ & $77.46$ & $76.82$ & $77.56$ \\
ATLOP$^{\star}$ & $\bm{h}_e$ ; $\bm{c}^{(s,o)}$ & $77.75 \pm 0.08$ & $78.41 \pm 0.10$ & $77.62$ & $78.38$ \\
GADePo (ours) & \texttt{<ent>} ; \texttt{<pent>} & $77.48 \pm 0.12$ & $78.19 \pm 0.14$ & $\textbf{77.70}$ & $\textbf{78.40}$ 
\vspace{-1ex}
\end{tabular} 
\caption{Results in percentage for the development and test sets of Re-DocRED. We report the results obtained by \citet{tan-etal-2022-revisiting} (ATLOP$^{\bullet}$) on Re-DocRED. ATLOP$^{\star}$ indicates our reimplementation of the previous method. We report the mean and standard deviation of Ign $F_1$ and $F_1$ on the development set, calculated from five training runs with distinct random seeds. We report the test score achieved by the best checkpoint on the development set. Ign $F_1$ refers to the $F_1$ score that excludes relational facts shared between the training and development/test sets.}
\label{tab:redocred_additional_results}
\end{table*}

\begin{table*}
\centering
\scalebox{0.9}{
    \begin{tabular}{l|l|ccc|ccc}
    \multicolumn{2}{c}{} & \multicolumn{3}{c}{\textbf{Dev}} & \multicolumn{3}{c}{\textbf{Test}} \\
    \textbf{Model} & \textbf{Aggregation} & $P$ & $R$ & $F_1$ & $P$ & $R$ & $F_1$ \\
    \hline
    ATLOP$^{\star}$ & $\bm{h}_e$ & $77.37 \pm 0.22$ & $77.40 \pm 0.31$ & $77.39 \pm 0.13$ & $\textbf{76.27}$ & $76.83$ & $76.55$ \\
    GADePo (ours) & \texttt{<ent>} & $72.96 \pm 0.96$ & $79.22 \pm 1.20$ & $75.96 \pm 0.99$ & $74.13$ & $\textbf{79.46}$ & $\textbf{76.70}$ \\
    \hline
    ATLOP$^{\diamond}$ & $\bm{h}_e$ ; $\bm{c}^{(s,o)}$ & $-$ & $-$ & $-$ & $77.89$ & $76.55$ & $77.21$ \\
    ATLOP$^{\star}$ & $\bm{h}_e$ ; $\bm{c}^{(s,o)}$ & $77.18 \pm 0.14$ & $77.98 \pm 0.66$ & $77.58 \pm 0.36 $ & $76.36$ & $78.86$ & $77.59$ \\
    GADePo (ours) & \texttt{<ent>} ; \texttt{<pent>} & $75.98 \pm 0.94$ & $80.54 \pm 0.72$ & $78.19 \pm 0.19$ & $\textbf{78.27}$ & $\textbf{79.03}$ & $\textbf{78.65}$ 
    \vspace{-1ex}
    \end{tabular}
}
\caption{Results in percentage for the development and test sets of HacRED. We report the results obtained by \citet{cheng-etal-2021-hacred} (ATLOP$^{\diamond}$) on HacRED. ATLOP$^{\star}$ indicates our reimplementation of the previous method. We report the mean and standard deviation of Precision ($P$), Recall ($R$) and $F_1$ on the development set, calculated from five training runs with distinct random seeds. We report the test score achieved by the best checkpoint on the development set.}
\label{tab:hacred_additional_results}
\end{table*}

\begin{table*}
\centering
\begin{tabular}{l|l|cc|cc}
\multicolumn{2}{c}{} & \multicolumn{2}{c}{\textbf{Dev}} & \multicolumn{2}{c}{\textbf{Test}} \\
\textbf{Model} & \textbf{Aggregation} & Ign $F_1$ & $F_1$ & Ign $F_1$ & $F_1$ \\
\hline
ATLOP$^{\star}$ & $\bm{h}_e$ & $59.66 \pm 0.20$ & $61.60 \pm 0.21$ & $59.22$ & $61.37$ \\
GADePo (ours) & \texttt{<ent>} & $59.04 \pm 0.52$ & $61.18 \pm 0.46$ & $\textbf{59.30}$ & $\textbf{61.63}$ \\
\hline
ATLOP$^{\circ}$ & $\bm{h}_e$ ; $\bm{c}^{(s,o)}$ & $61.32 \pm 0.14$ & $63.18 \pm 0.19$ & $ 61.39$ & $63.40$ \\
ATLOP$^{\star}$ & $\bm{h}_e$ ; $\bm{c}^{(s,o)}$ & $61.41 \pm 0.26$ & $63.38 \pm 0.28$ & $\textbf{61.62}$ & $63.72$ \\
GADePo (ours) & \texttt{<ent>} ; \texttt{<pent>} & $61.19 \pm 0.55$ & $63.26 \pm 0.48$ & $61.52$ & $\textbf{63.75}$ 
\vspace{-1ex}
\end{tabular} 
\caption{Results in percentage for the development and test sets of DocRED. We report the results obtained by \citet{zhou2021atlop} (ATLOP$^{\circ}$) on DocRED. ATLOP$^{\star}$ indicates our reimplementation of the previous method. We report the mean and standard deviation of Ign $F_1$ and $F_1$ on the development set, calculated from five training runs with distinct random seeds. We report the test score achieved by the best checkpoint on the development set. Ign $F_1$ refers to the $F_1$ score that excludes relational facts shared between the training and development/test sets.}
\label{tab:docred_results}
\end{table*}

\subsection{Additional Results}
\label{subsec:additional_results}

\paragraph{Re-DocRED and HacRED} Table \ref{tab:redocred_additional_results} and Table \ref{tab:hacred_additional_results} present additional results for Re-DocRED and HacRED, respectively. In addition to the results outlined in Section \ref{sec:experiments}, these tables include the mean and standard deviation on the development set, calculated from five training runs with distinct random seeds, as reported in Appendix Subsection \ref{subsec:training_details}.

\begin{figure}[h]
    \centering
    \centering
    \includegraphics[width=\linewidth]{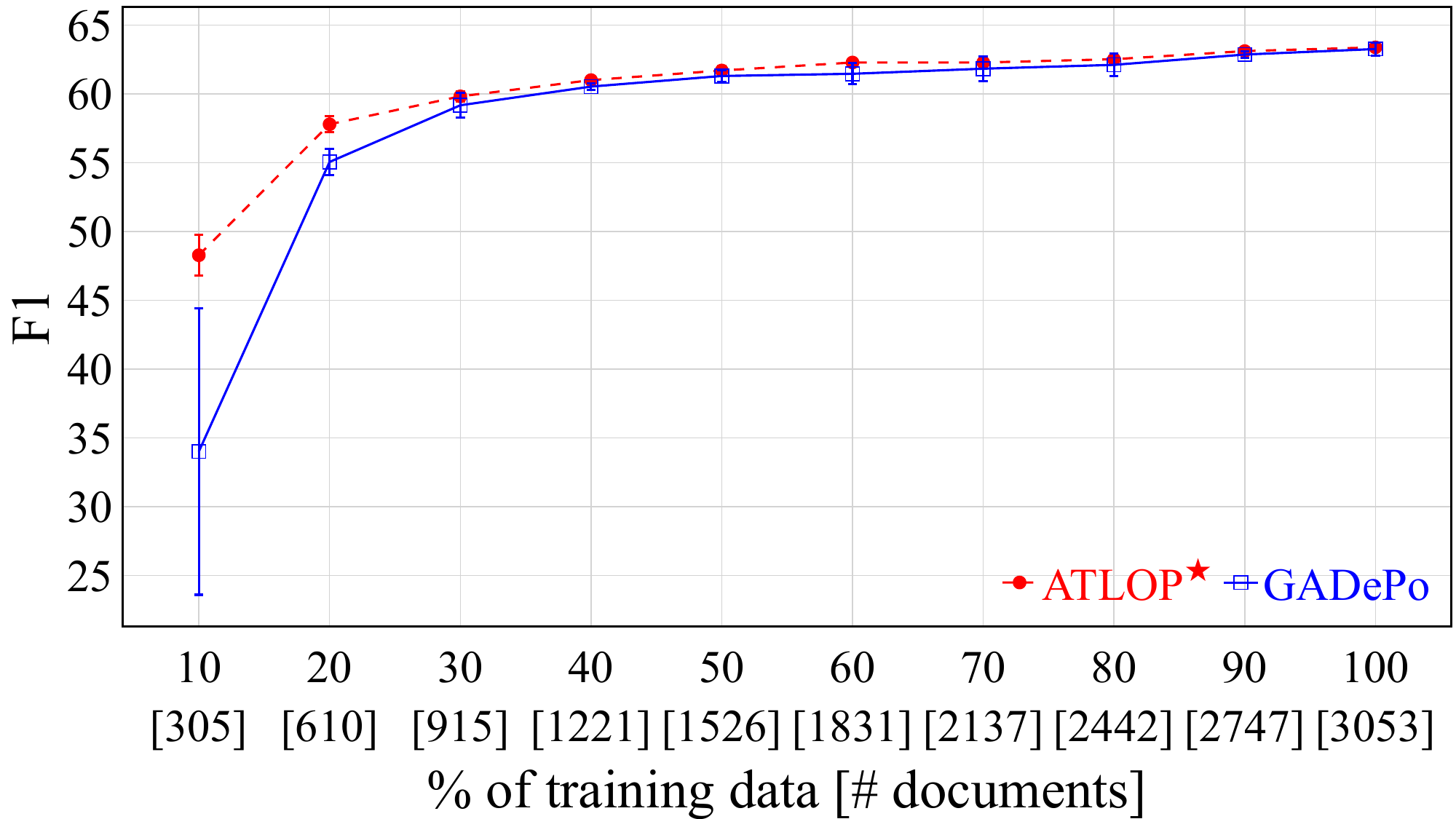}
    \vspace{-1ex}\\
    \caption{Performance of ATLOP$^{\star}$ ($\bm{h}_e$ ; $\bm{c}^{(s,o)}$) and GADePo (\texttt{<ent>} ; \texttt{<pent>}) on the development set under varying data availability conditions on DocRED. The $x$-axis represents the percentage and number of documents from the training dataset, while the $y$-axis displays the $F_1$ score in percentage. Each point on the graph represents the mean value, while error bars indicate the standard deviation derived from five distinct training runs with separate random seeds.}
    \label{fig:docred_data_ablation_study}
\end{figure}

\paragraph{DocRED results} The DocRED \cite{yao-etal-2019-docred} dataset consists of $56,354$ facts, $96$ relations, $5,053$ documents, and $26.2$ average number of entities per document. In line with the approach taken for Re-DocRED and HacRED, Table \ref{tab:docred_results} and Figure \ref{fig:docred_data_ablation_study} illustrate the results for DocRED.

\end{document}